%% file: main.tex
\documentclass[10pt,twocolumn,letterpaper]{article}

\usepackage{cvpr}              % To produce the CAMERA-READY version
\usepackage{cancel}

\usepackage[table]{xcolor} 
\definecolor{darkgreen}{RGB}{0,128,0}
\definecolor{darkblue}{RGB}{70,130,230}
\definecolor{lightgreen}{RGB}{220,255,220}
\definecolor{cvprblue}{rgb}{0.21,0.49,0.74}
\usepackage[pagebackref,breaklinks,colorlinks,allcolors=cvprblue]{hyperref}

\usepackage{multirow, makecell, booktabs, pifont, tabularx, array, threeparttable}

\def\paperID{6734} % *** Enter the Paper ID here
\def\confName{CVPR}
\def\confYear{2026}

\title{Bridge: Basis-Driven Causal Inference Marries VFMs for Domain Generalization}

\author{
Mingbo Hong$^{1}$ \quad
Feng Liu$^{2}$ \quad
Caroline Gevaert$^{1}$ \quad
George Vosselman$^{1}$ \quad
Hao Cheng$^{1}$\\[0.5em]
$^{1}$University of Twente \quad
$^{2}$Drexel University\\[0.5em]
{\tt\small mingbohong97@gmail.com, fl397@drexel.edu, c.m.gevaert@utwente.nl}\\
{\tt\small george.vosselman@utwente.nl, h.cheng-2@utwente.nl}
}

\begin{document}
\maketitle
\input{sec/0_abstract}    
\input{sec/1_intro}

\input{sec/2_related_work}
\input{sec/3_method}

\input{sec/4_experiments}
\input{sec/5_conclusion}

\section*{Acknowledgments}
{\raggedright
This work used the Dutch national e-infrastructure with the support of the SURF Cooperative using grant no.~EINF-\\15118.\par
}
{
    \small
    \bibliographystyle{ieeenat_fullname}
    \bibliography{main}
}

% WARNING: do not forget to delete the supplementary pages from your submission 
% \input{sec/X_suppl}

\end{document}

%% file: sec/0_abstract.tex
\begin{abstract}
Detectors often suffer from degraded performance, primarily due to the distributional gap between the source and target domains. 
This issue is especially evident in single-source domains with limited data, as models tend to rely on confounders (e.g., illumination, co-occurrence, and style) from the source domain, leading to spurious correlations that hinder generalization. 
To this end, this paper proposes a novel Basis-driven framework for domain generalization, namely \textbf{\textit{Bridge}}, that incorporates causal inference into object detection. 
By learning the low-rank bases for front-door adjustment, \textbf{\textit{Bridge}} blocks confounders' effects to mitigate spurious correlations, while simultaneously refining representations by filtering redundant and task-irrelevant components.
\textbf{\textit{Bridge}} can be seamlessly integrated with both discriminative (e.g., DINOv2/3, SAM) and generative (e.g., Stable Diffusion) Vision Foundation Models (VFMs).
Extensive experiments across multiple domain generalization object detection datasets, i.e., Cross-Camera, Adverse Weather, Real-to-Artistic, Diverse Weather Datasets, and Diverse Weather DroneVehicle (our newly augmented real-world UAV-based benchmark), underscore the superiority of our proposed method over previous state-of-the-art approaches.
The project page is available at:
\url{https://mingbohong.github.io/Bridge/}.

\vspace{-5mm}
\end{abstract}

%% file: sec/1_intro.tex
\section{Introduction}
\label{sec:intro}

% Recent advances in deep learning have significantly improved object detection. However, these data-driven models can inherit or amplify biases from training datasets, leading to poor transferability across domains~\cite{mehrabi2021survey}.
% To address these challenges, researchers have explored various techniques [ref] to enhance the robustness and generalization capabilities of deep learning models. 
% One promising technique is domain generalization~\cite{blanchard2011generalizing} (DG),
% which aims to learn a model from one or multiple source domains that can perform well on out-of-distribution (OOD) target domains. 
% Existing mainstream DG methods primarily focus on learning domain-invariant representations~\cite{li2018deep,rahman2020correlation,albuquerque2019generalizing,deng2020representation}, expanding source distributions through data augmentation~\cite{tobin2017domain,peng2018sim,xu2021fourier,zhou2021domain}, or leveraging the strong prior knowledge of Vision Foundation Models (VFMs) ~\cite{wei2024stronger,yun2025soma,he2025generalized,he2025boosting}.

% Recent breakthroughs in deep learning have substantially advanced the accuracy and efficiency of object detection systems across a wide range of applications. 
% Despite these achievements, 
Despite recent advances, data-driven object detection models often inherit or amplify dataset biases, limiting their cross-domain transferability~\cite{mehrabi2021survey}. 
They frequently perform poorly across diverse viewpoints and imaging conditions from ground-level scenes to aerial imagery, which hinders generalization.
% For instance, detectors trained on standard datasets may struggle when applied to remote sensing imagery, such as UAV images captured under different weather conditions, where variations in illumination and scene appearance can dramatically alter object visibility. 
\begin{figure}[t]
    \centering
    \includegraphics[width=0.47\textwidth]{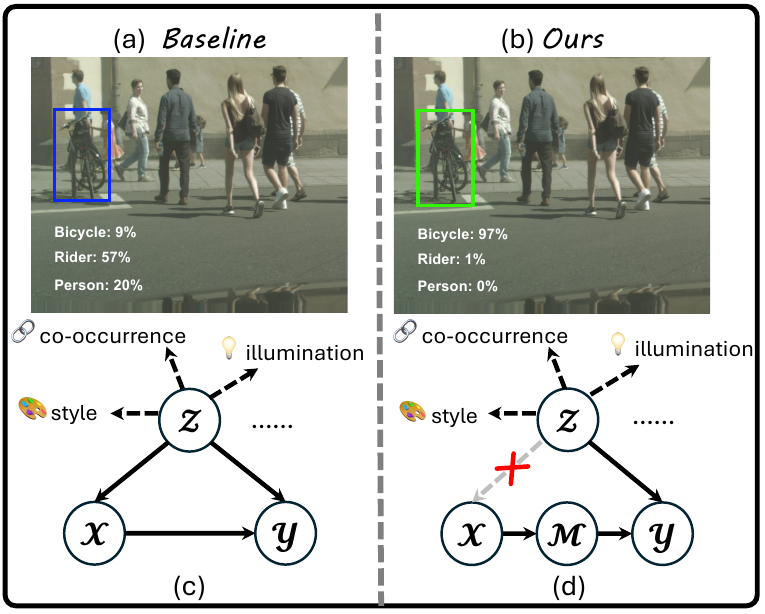}
    \caption{
(a) The baseline model, using a frozen DINOv2~\cite{oquab2023dinov2} backbone, misclassifies the object due to spurious correlations caused by the confounder: co-occurrence of a bicycle with nearby pedestrians.
(b) Our method correctly identifies the bicycle by blocking these spurious correlations.
(c) Models may rely on spurious correlations between the input $\mathcal{X}$ and the label $\mathcal{Y}$ induced by the confounder $\mathcal{Z}$.
% (c) Models may rely on spurious correlations between the input $\mathcal{X}$ and the label $\mathcal{Y}$ through confounders $\mathcal{Z}$.
% (d) In contrast, our method implements a front-door adjustment mechanism, where the mediating variable $\mathcal{M}$ mitigates the influence of spurious correlations induced by $\mathcal{Z}$, thereby enabling the model to focus on the features that truly determine $\mathcal{Y}$.
(d) In contrast, our method uses a front-door adjustment mechanism, where the mediating variable $\mathcal{M}$ mitigates spurious correlations induced by $\mathcal{Z}$ and enables the model to focus on the features that truly determine $\mathcal{Y}$.
}
    \label{fig:teaser}
    \vspace{-6mm}
\end{figure}

To address these limitations, domain generalization (DG)~\cite{blanchard2011generalizing} has emerged as a particularly promising paradigm, aiming to train models on one or multiple source domains to generalize effectively to unseen out-of-distribution (OOD) target domains. 
Existing DG approaches primarily focus on learning domain-invariant representations~\cite{li2018deep,rahman2020correlation,albuquerque2019generalizing,deng2020representation}, expanding source distributions through data augmentation~\cite{tobin2017domain,peng2018sim,fact,zhou2021domain}, or leveraging the strong prior knowledge embedded in Vision Foundation Models (VFMs)~\cite{wei2024stronger,yun2025soma,he2025generalized,he2025boosting}.

However, many DG approaches often neglect the confounding effects that arise from limited single-domain datasets. 
These confounders (denoted as $\mathcal{Z}$), including variations in illumination, co-occurrence patterns, or stylistic discrepancies, can bias the learned representations ($\mathcal{X}$) and create spurious correlations with object labels ($\mathcal{Y}$), see Fig.~\ref{fig:teaser}.
Such biases distort the causal link between visual features and labels~\cite{wang2020visual,zhang2022multiple,zhang2020causal,shao2021improving}, degrading out-of-distribution generalization.
For instance, a detector with a frozen DINOv2~\cite{oquab2023dinov2} backbone pretrained on \textit{142M} images but fine-tuned on only \textit{3,000} Cityscapes samples~\cite{cordts2016cityscapes} may misclassify a bicycle near a person as a rider (57\%) or even as a person (20\%) due to spatial correlations (Fig.~\ref{fig:teaser}~(a)).
The backbone offers substantial representational strength, yet overlooking spurious correlations in downstream tasks risks wasting much of this capacity.
Hence, to mitigate these effects, it is crucial to uncover and model the causal mechanisms behind the data, guiding the model to focus on causally relevant features rather than spurious correlations.

In this paper, we introduce \textbf{\textit{Bridge}}, a \textbf{B}asis-d\textbf{ri}ven framework for \textbf{D}omain \textbf{Ge}neralization that incorporates causal inference~\cite{pearl2016causal,neuberg2003causality} into object detection. 
% By uncovering intrinsic causal structures, 
By blocking the adverse effects of confounders,
\textbf{\textit{Bridge}} enables the model to mitigate the influence of confounders.
% Specifically, we propose a novel Causal Basis Block (CBB) that learns a front-door adjustment from a basis-learning perspective to mitigate the adverse effects of confounders~\cite{pearl2018book}.
Specifically, we propose a novel Causal Basis Block (CBB) that is designed in accordance with the front-door adjustment~\cite{pearl2018book}, formulated from a basis-learning perspective to mitigate the adverse effects of confounders.
Unlike prior works~\cite{wang2024vision,liang2024confounded,zhang2022multiple,zhang2020causal} that depend on external confounder dictionaries or involve elaborate post-processing mechanisms (e.g., clustering~\cite{zhang2020causal}, momentum updates~\cite{liang2024confounded}) with limited flexibility and scalability,
our CBB is proposed as an end-to-end, plug-and-play module tailored for VFMs-based object detection frameworks, allowing seamless integration while maintaining high flexibility and scalability.
% our CBB is end-to-end and can be seamlessly integrated into existing object detection frameworks via plug-and-play.
In our design, we build the Domain Generalization Object Detection (DGOD) framework upon frozen VFMs by integrating the CBB to perform causal modeling, avoiding costly fine-tuning the whole network. 
By blocking spurious correlations (Fig.~\ref{fig:teaser}~(d)), the framework allows the model to focus on inherent causality, yielding more robust and generalizable representations for downstream tasks and improving performance (Fig.~\ref{fig:teaser}~(b)).

Furthermore, we note that commonly used DGOD benchmarks~\cite{cordts2016cityscapes,yu2020bdd100k,sakaridis2018semantic,everingham2010pascal,inoue2018cross,wu2022single} lack diverse aerial viewpoints and imaging conditions.
For a more comprehensive evaluation, we extend the DroneVehicle dataset~\cite{sun2022drone} by annotating weather conditions and organizing the images into four scenarios: Clear, Dark, Foggy, and Extreme Dark.

We summarize our main contributions as follows:
\begin{itemize}
\item We propose \textbf{\textit{Bridge}}, a novel framework that can be seamlessly integrated with both discriminative (e.g., DINOv2~\cite{oquab2023dinov2}, DINOv3~\cite{simeoni2025dinov3}, SAM~\cite{kirillov2023segment}) and generative (e.g., Stable Diffusion~\cite{rombach2022high}) Vision Foundation Models (VFMs), aiming to mitigate spurious correlations in downstream domain generalization object detection.
\item We introduce a novel Causal Basis Block that constructs front-door adjustments from a basis-learning perspective, which can be trained end-to-end and improves model generalization by producing robust representations.
\item We conduct extensive experiments on five DGOD benchmarks. 
\textbf{\textit{Bridge}} surpasses the previous runner-up across all the benchmarks by 
$\mathbf{+3.8}$, $\mathbf{+2.9}$, $\mathbf{+2.4}$, $\mathbf{+0.4}$, 
and $\mathbf{+1.5}$ in mAP, and consistently improves the VFMs (SAM, DINOv2, DINOv3) with average gains of 
$\mathbf{+2.7}$, $\mathbf{+4.5}$, $\mathbf{+3.1}$, $\mathbf{+1.8}$, 
and $\mathbf{+1.8}$ in mAP, respectively.

% \item We conduct extensive experiments on five domain generalization benchmarks, demonstrating that \textbf{\textit{Bridge}} consistently enhances robustness and generalization across both discriminative (SAM, DINOv2, DINOv3) and generative (Stable Diffusion) Vision Foundation Models, achieving average mAP gains of $\mathbf{+3.0}$, $\mathbf{+4.1}$, $\mathbf{+2.8}$, $\mathbf{+1.4}$, and $\mathbf{+1.8}$.
% Noteworthily, among the five benchmarks, we enrich an existing UAV object detection dataset with manually annotated weather condition labels, establishing a much-needed benchmark for evaluating domain generalization under diverse weather scenarios in the remote sensing community.
\item We enrich an existing UAV object detection dataset with manually annotated weather condition labels, establishing a much-needed benchmark for evaluating domain generalization under diverse weather scenarios in the remote sensing community.

\end{itemize}

%% file: sec/2_related_work.tex
\section{Related Work}
\label{sec:related-works}

\paragraph{Domain Generalization for Visual Perception.}
% As a crucial yet challenging task, domain generalization~\cite{blanchard2011generalizing} (DG) aims to fully exploit the limited knowledge from single or multiple source domains to generalize well to out-of-distribution (OOD) target domains. 
Domain generalization~\cite{blanchard2011generalizing} has been widely studied in various visual perception tasks, 
including image classification~\cite{li2017deeper,li2019feature}, object detection~\cite{malisiewicz2011ensemble,hoffman2016learning}, medical imaging~\cite{liu2020ms,liu2020shape}, semantic segmentation~\cite{yue2019domain,volpi2019addressing}, and beyond. 
To tackle the DG problem, a wide range of approaches have been proposed from different perspectives, 
including data augmentation~\cite{volpi2019addressing, shi2020towards}, adversarial training~\cite{li2018deep,rahman2020correlation,albuquerque2019generalizing}, disentangled representation learning~\cite{li2017deeper,khosla2012undoing}, meta-learning~\cite{li2018learning,balaji2018metareg}, and more.
More recently, several works have explored the potential of Vision Foundation Models (VFMs)~\cite{caron2021emerging,kirillov2023segment} 
in various DG tasks~\cite{he2025generalized,wei2024stronger,yun2025soma}.

Building on VFMs, our \textbf{\textit{Bridge }} aims to mitigate spurious correlations that may arise in the process of leveraging frozen VFMs for downstream training, 
especially when trained on a single, data-scarce source domain.
\vspace{-1mm}
\paragraph{Vision Foundation Models (VFMs).}
VFMs have rapidly emerged in recent years, driven by large-scale training data in self-supervised or semi-supervised settings.
ViT~\cite{dosovitskiy2020image} pioneered the use of large-scale transformer architectures for image classification, demonstrating the scalability of transformer-based models on visual tasks.
Building on this, transformer-based self-supervised paradigms, such as DINO~\cite{caron2021emerging,oquab2023dinov2,simeoni2025dinov3} and masked autoencoders~\cite{he2022masked} (MAE), have been proposed to learn robust image representations for downstream tasks in a label-efficient manner.
Meanwhile, models such as the Segment Anything Model~\cite{kirillov2023segment,ravi2024sam2} (SAM) extend the foundation model concept to dense prediction tasks by providing flexible multimodal prompts for diverse segmentation scenarios.
Interestingly, generative models such as Stable Diffusion~\cite{rombach2022high} have also been explored as foundation models, leveraging large-scale pretraining to support visual perception tasks~\cite{he2025generalized,he2024diffusion,he2025boosting}.
% VFMs have rapidly emerged in the vision community in recent years, benefiting from large-scale training data in a self-supervised or semi-supervised manner. 
% Amongst others, ViT~\cite{dosovitskiy2020image} pioneered the application of large-scale transformer architectures to image classification, demonstrating the scalability of transformer-based models on visual tasks. 
% Building on this, transformer-based self-supervised paradigms, such as DINO~\cite{caron2021emerging,oquab2023dinov2,simeoni2025dinov3} and masked autoencoders~\cite{he2022masked} (MAE), have been proposed to learn robust image representations for downstream tasks in a label-efficient manner.
% Meanwhile, models such as the Segment Anything Model~\cite{kirillov2023segment,ravi2024sam2} (SAM) extend the foundation model concept to dense prediction tasks, providing flexible multimodal prompts to accommodate diverse segmentation scenarios. 
% Interestingly, generative models, such as Stable Diffusion~\cite{rombach2022high}, have also been explored as a foundation model, leveraging large-scale pretraining to support visual perception tasks ~\cite{he2025generalized,he2024diffusion,he2025boosting}.

Despite VFMs' remarkable performance in downstream tasks, most existing methods directly adopt these models while overlooking spurious correlations, hindering their performance, particularly in domain generalization tasks.
\vspace{-2mm}
\paragraph{Causal Inference.}
Causal inference~\cite{pearl2016causal,neuberg2003causality} has been widely used in deep learning to mitigate spurious correlations, with applications to various vision tasks, including image classification~\cite{chalupka2014visual,lopez2017discovering}, visual question answering~\cite{liu2023cross,yang2021deconfounded,wang2020visual}, semantic segmentation~\cite{zhang2020causal,shao2021improving}, and object detection~\cite{zhang2022multiple}.
Existing methods typically perform causal intervention through front-door or back-door adjustment. For example, Wang et al.~\cite{wang2024vision} integrate both front-door and back-door adjustment in vision-and-language navigation, while Zhang et al.~\cite{zhang2022multiple} develop a causal framework for object detection under adverse weather based on back-door adjustment with explicitly modeled confounders. 

Different from such methods that rely on explicit confounder construction, we realize front-door adjustment through basis learning, without manually specifying confounders or introducing external confounder dictionaries and task-specific post-processing. Moreover, the learned bases naturally induce a low-rank structure, preserving causal representations in a compact form and improving flexibility and scalability.
\vspace{-1mm}

% In contrast, our method formulates front-door adjustment through basis learning, avoiding the need for explicitly specified confounders and reducing reliance on task-specific post-processing, while simultaneously incorporating a low-rank structure to preserve causal representations in a compact form.

%% file: sec/3_method.tex
\section{Method}
\label{sec:method}

\begin{figure*}[t]
    \centering
    \includegraphics[width=1.0\textwidth]{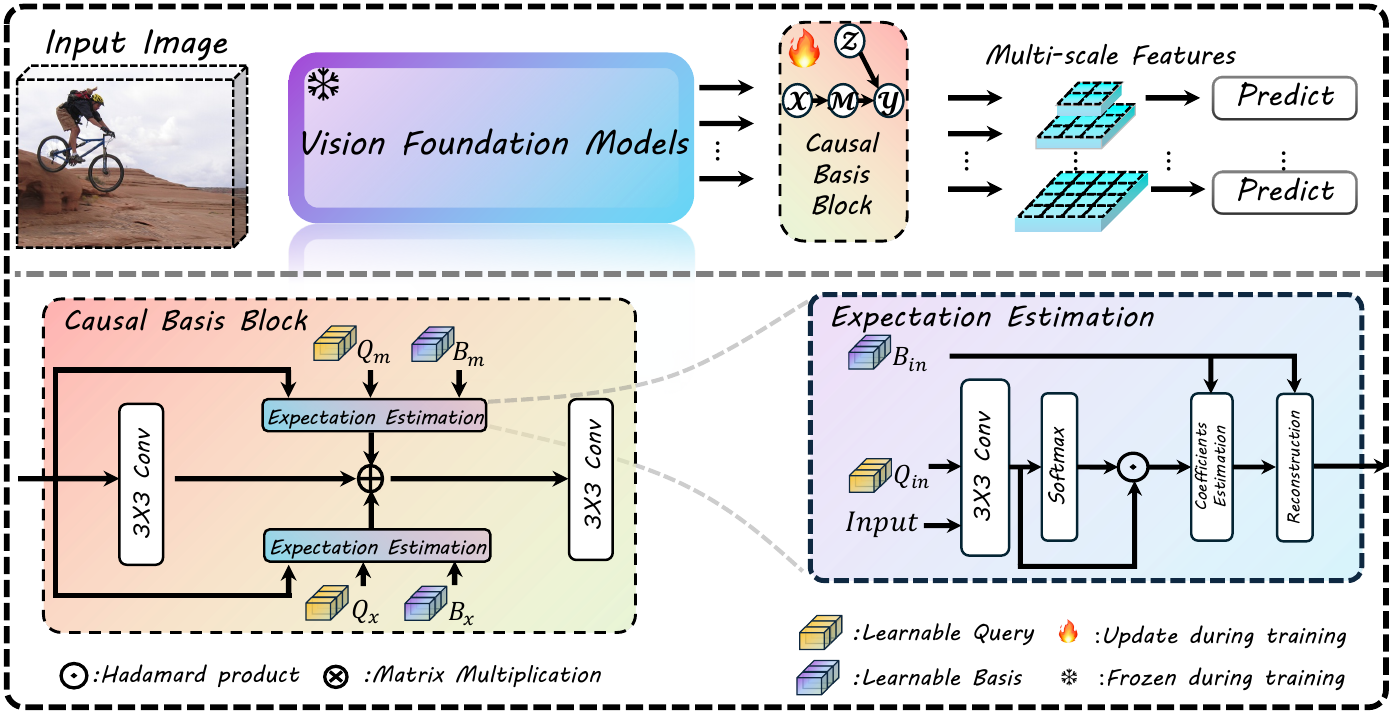}
    % \caption{Overview of the proposed \textbf{\textit{Bridge}}. Multi-scale features are first extracted from the Vision Foundation Model, then calibrated using Causal Basis Blocks，and finally fed into task-specific heads for prediction.}
    \caption{Overview of the proposed \textbf{\textit{Bridge}}. Multi-scale features are first extracted from the Vision Foundation Model, then calibrated using the Causal Basis Block (CBB). Within the CBB, the Coefficients Estimation and Reconstruction components in Expectation Estimation refer to Equation~\ref{equ:coefficients} and Equation~\ref{equ:expected_reconstruction}, respectively. The calibrated features are finally fed into task-specific heads for prediction.}
    \label{fig:pipe}
\end{figure*}

\subsection{Preliminaries}
\subsubsection{Task Formulation}
In DG, given source data $\mathcal{D}_s=\{\mathbf{x}^s,\mathbf{y}^s\}$, the goal is to learn a model $f:\mathcal{X}\rightarrow\mathcal{Y}$ that generalizes to an unseen target domain $\mathcal{D}_t=\{\mathbf{x}^t\}$. Here, $\mathcal{X}$ and $\mathcal{Y}$ denote the input and output spaces. The marginal distributions differ across domains, i.e., $\mathcal{P}_X^{s}\neq\mathcal{P}_X^{t}$, mainly due to discrepancies in data sources.
\subsubsection{Front-door Adjustment}
Visual recognition involves various confounding factors, such as background, illumination, texture, and context, which bias the model’s estimation of the underlying relationship between inputs and semantic labels. 
To illustrate this effect, by Bayes' rule, we have
\begin{equation}
    \mathcal{P}(\mathcal{Y} \mid \mathcal{X}) 
    = \sum_{\mathcal{Z}} \mathcal{P}(\mathcal{Y} \mid \mathcal{X}, \mathcal{Z}) \, \mathcal{P}(\mathcal{Z} \mid \mathcal{X}),
    \label{equ:bayes_rule}
\end{equation}
where $\mathcal{Z}$ denotes confounders that jointly affect $\mathcal{X}$ and $\mathcal{Y}$, inducing spurious correlations. 
Note that $\mathcal{P}(\mathcal{Z}\mid\mathcal{X})$ may \textbf{dominate} the prediction of the model when samples are limited, leading the model to rely on spurious correlations rather than causal features (please refer to the supplementary material for further explanation). 
To block the back-door paths $\mathcal{X} \leftarrow \mathcal{Z} \rightarrow \mathcal{Y}$, the back-door adjustment using the do-operator~\cite{pearl2018book} conditions on $\mathcal{Z}$ as follows:
\begin{equation}
    \mathcal{P}(\mathcal{Y} \mid \mathrm{do}(\mathcal{X})) 
    = \sum_{\mathcal{Z}} \mathcal{P}(\mathcal{Y} \mid \mathcal{X}, \mathcal{Z}) \, \mathcal{P}(\mathcal{Z}),
    \label{equ:back_door_adjustment}
\end{equation}

However, in practice, some confounders $\mathcal{Z}$ may be unobserved or difficult to measure, making such an adjustment infeasible. 
Fortunately, front-door adjustment provides a principled way to mitigate such spurious correlation effects without explicitly modeling the confounders. 
Specifically, an observed mediator $\mathcal{M}$ on the causal path from $\mathcal{X}$ to $\mathcal{Y}$ enables the causal effect of $\mathcal{X}$ on $\mathcal{Y}$ to be identified.
The front-door adjustment is given by:
% Mathematically, the front-door adjustment is formulated as follows:
\begin{equation}
\resizebox{0.433\textwidth}{!}{%
$\displaystyle
\begin{aligned}
\mathcal{P}(\mathcal{Y} \mid \mathrm{do}(\mathcal{X})) 
&= \sum_{\mathcal{M}} \mathcal{P}(\mathcal{Y} \mid \mathcal{M}, \mathrm{do}(\mathcal{X})) 
   \, \mathcal{P}(\mathcal{M} \mid \mathrm{do}(\mathcal{X})) \\
&= \sum_{\mathcal{M}} \mathcal{P}(\mathcal{M} \mid \mathcal{X}) 
   \sum_{\mathcal{X}'} \mathcal{P}(\mathcal{Y} \mid \mathcal{X}', \mathcal{M}) \, \mathcal{P}(\mathcal{X}') \\
&= \mathbb{E}_{\mathcal{M} \sim \mathcal{P}(\mathcal{M} \mid \mathcal{X})} 
   \Big[ \, \mathbb{E}_{\mathcal{X}' \sim \mathcal{P}(\mathcal{X})} 
        [\, \mathcal{P}(\mathcal{Y} \mid \mathcal{X}', \mathcal{M}) \,] 
   \Big],
\end{aligned}%
$}
\label{equ:front_door_adjustment}
\end{equation}
where $\mathcal{X}'$ denotes values of $\mathcal{X}$ sampled from its marginal distribution $\mathcal{P}(\mathcal{X})$, and to avoid explicitly computing the full nested expectations over all possible $\mathcal{X}'$ and $\mathcal{M}$ during training, we employ the Normalized Weighted Geometric Mean (NWGM)~\cite{xu2015show} to approximate Eq.~(\ref{equ:front_door_adjustment}) as follows:
\begin{equation}
\resizebox{0.43\textwidth}{!}{$
\mathbb{E}_{\mathcal{M} \sim \mathcal{P}(\mathcal{M} \mid \mathcal{X})}
\Big[
\mathbb{E}_{\mathcal{X}' \sim \mathcal{P}(\mathcal{X})}
[\mathcal{P}(\mathcal{Y}\mid \mathcal{X}',\mathcal{M})]
\Big]
\overset{\text{NWGM}}{\approx}
\mathcal{P}\Big(
\mathcal{Y}\mid
\mathbb{E}_{\mathcal{X}' \sim \mathcal{P}(\mathcal{X})}[\mathcal{X}'],
\mathbb{E}_{\mathcal{M} \sim \mathcal{P}(\mathcal{M}\mid \mathcal{X})}[\mathcal{M}]
\Big)
$}
\label{equ:nwgm_approximation}
\end{equation}

Recently, Wang et al.~\cite{wang2024vision} extended the front-door adjustment from the network's final prediction layer to intermediate feature layers by formulating it as a linear mapping model. Accordingly, Eq.~(\ref{equ:nwgm_approximation}) can be further reformulated as:
% \begin{equation}
% \mathcal{F}(\mathcal{X})
% =
% \mathbb{E}_{\mathcal{M} \sim \mathcal{P}(\mathcal{M} \mid \mathcal{X})}[\mathcal{M}] 
% +
% \mathbb{E}_{\mathcal{X}' \sim \mathcal{P}(\mathcal{X})}[\mathcal{X}'].
% \label{equ:front_door_adjustment_linear}
% \end{equation}
\begin{equation}
\resizebox{0.4\textwidth}{!}{$
\mathcal{P}(\mathcal{Y}\mid do(\mathcal{X}))
\approx
\mathcal{P}\Big(
\mathcal{Y}\mid
\mathbb{E}_{\mathcal{X}' \sim \mathcal{P}(\mathcal{X})}[\mathcal{X}']
+
\mathbb{E}_{\mathcal{M} \sim \mathcal{P}(\mathcal{M} \mid \mathcal{X})}[\mathcal{M}]
\Big)
$}
\label{equ:front_door_adjustment_linear}
\end{equation}
Thus, implementing the front-door adjustment reduces to estimating the two expectations 
$\mathbb{E}_{\mathcal{M} \mid \mathcal{X}}[\mathcal{M}]$ and $\mathbb{E}_{\mathcal{X}'}[\mathcal{X}']$.
\subsection{Overview}
This section presents \textbf{\textit{Bridge}}, a DGOD method built on VFMs. \textbf{\textit{Bridge}} applies front-door adjustment to VFMs' features through the basis learning mechanism, producing causal and robust representations for downstream tasks.
Fig.~\ref{fig:pipe} illustrates the overall architecture of \textbf{\textit{Bridge}}. Specifically, given an input image, multi-scale feature maps are first extracted using the VFMs, then refined through the proposed Causal Basis Block to improve robustness and generalization.
The resulting calibrated features are finally passed to task-specific heads for prediction.
\subsection{Causal Basis Block}
Inspired by dictionary learning~\cite{tovsic2011dictionary}, in which data are represented as combinations of basis vectors, 
we approximate the expectations in Equation~\ref{equ:front_door_adjustment_linear} 
as linear combinations of learnable basis vectors:
\begin{equation}
    \mathbb{E}[\mathcal{V}] \approx \frac{1}{S} \sum_{i=1}^S \sum_{k=1}^K c_{ik} b_k,
    \label{equ:expectation_basis_avg}
\end{equation}
where $\mathcal{V}$ denotes a generic variable (e.g., $\mathcal{M}\mid\mathcal{X}$ or $\mathcal{X}'$), 
$b_k$ ($k=1,\dots,K$) are the learnable basis vectors, 
$S$ is the total number of samples, and $c_{ik}$ are the sample coefficients.

\subsubsection{Expectations Estimation} 
\label{sec:coefficients_estimation}
To implement Equation~\ref{equ:expectation_basis_avg}, we describe how the basis vectors and their coefficients are learned.
We design a module termed the \textbf{Causal Basis Block} (CBB), which jointly estimates coefficients and learns bases in an end-to-end differentiable manner.

Given input feature maps $\mathcal{X}_\mathrm{in} \in \mathbb{R}^{B \times N \times C}$ (where $N = H \times W$, with $H$ and $W$ denoting the height and width), the CBB introduces a set of learnable \textit{Sample Queries}
$\mathcal{Q}_s \in \mathbb{R}^{S \times C}$, which act similarly to the 
\textit{object queries} in DETR~\cite{carion2020end} and Mask2Former~\cite{cheng2022masked}.
During training, these queries implicitly aggregate global representations from the training set, guiding the estimation of expectations. Specifically, we use the \textit{Sample Queries} $\mathcal{Q}_s$ to approximate the expectation of the input features:
\begin{equation}
\resizebox{0.43\textwidth}{!}{$
\mathcal{X}'_q = \mathcal{X}_{in}\mathcal{Q}_s^{\top}, \quad
p = \operatorname{Softmax}(\mathcal{X}'_q), \quad
\mathcal{A} = \sum_{i=1}^{S} p_i\,\mathcal{X}'_{q,i}
$}
\label{equ:avg_correlation_corrected}
\end{equation}
where $\mathcal{X}'_q \in \mathbb{R}^{B \times N \times S}$ denotes the responses of the \textit{Sample Queries} to $\mathcal{X}_{in}$.
The Softmax is applied along the sample dimension $S$ to obtain
$p \in \mathbb{R}^{B \times N \times S}$.
The expected spatial weighting map $\mathcal{A} \in \mathbb{R}^{B \times N \times 1}$ is obtained by a weighted sum of the query responses $\mathcal{X}'_{q,i}$ with weights $p_i$.

We then obtain a query-guided representation by reweighting the input features with $\mathcal{A}$:
\begin{equation}
    \mathcal{X}_{q} = \mathcal{A}  \odot \mathcal{X}_\mathrm{in} \in \mathbb{R}^{B \times N \times C},
    \label{equ:query_aggregation}
\end{equation}
where $\mathcal{X}_q$ emphasizes the generalized information in $\mathcal{X}_\mathrm{in}$ through the spatial weighting map $\mathcal{A}$.

With the query-guided features $\mathcal{X}_{q}$, we introduce a set of learnable bases
$\mathcal{B} = [b_1, \dots, b_K] \in \mathbb{R}^{K \times C}$ ($K < C$), providing a low-rank subspace for representation.
Projecting $\mathcal{X}_{q}$ onto this subspace enables reconstruction with fewer bases while preserving the most representative components. The coefficients $\mathcal{C}$ of $\mathcal{X}_{q}$ w.r.t.\ $\mathcal{B}$ are computed as:
\begin{equation}
   \mathcal{C} = \mathcal{X}_{q}\mathcal{B}^{\top}(\mathcal{B}\mathcal{B}^{\top})^{-1}
   \in \mathbb{R}^{B \times N \times K},
   \label{equ:coefficients}
\end{equation}
where $(\mathcal{B}\mathcal{B}^{\top})^{-1}$ serves as a normalization term since the bases may not remain orthogonal during training.

By first reweighting with $\mathcal{A}$ to highlight generalizable information and then projecting onto the low-rank subspace to remove redundancy, the estimated coefficients $\mathcal{C}$ are encouraged to approximate $\frac{1}{S}\sum_{i=1}^S c_{i}$, capturing the intrinsic representation aggregated by the \textit{Sample Queries} and filtered through the subspace projection.
We finally estimate the expectation in Equation~\ref{equ:expectation_basis_avg} as:
\begin{equation}
\mathbb{E}[\mathcal{X}_\mathrm{in}] \approx \mathcal{C} \mathcal{B}\ \in \mathbb{R}^{B \times N \times C}.
\label{equ:expected_reconstruction}
\end{equation}

In summary, since closed-form expectations in complex representation spaces are intractable, 
we employ \textit{Sample Queries} and subspace projection to approximate them: the queries aggregate generalizable cross-sample information, and the projection filters it into a low-rank space spanned by $\mathcal{B}$, rejecting redundancy while preserving representative structure (illustratively: $\mathbb{R}^{N \times C} \to \mathbb{R}^{N \times K} \to \mathbb{R}^{N \times C}$).
The low-rank, learnable bases naturally capture the principal modes of the feature space, and this compression effectively aligns features with the core directions of the sample distribution, thereby approximating the sample expectation.

% Note that the entire procedure is \textbf{differentiable} and can be optimized end-to-end with the network.
% At inference, $\mathcal{B}^{\top}(\mathcal{B}\mathcal{B}^{\top})^{-1}\mathcal{B}$ can be precomputed and absorbed into a single $C \times C$ matrix, reducing computation and simplifying deployment.

Note that the entire procedure is \textbf{differentiable} and can be optimized end-to-end.
At inference, $\mathcal{B}^{\top}(\mathcal{B}\mathcal{B}^{\top})^{-1}\mathcal{B}$ can be precomputed as a fixed $C \times C$ matrix, reducing computation and simplifying deployment.

\subsubsection{Feature Aggregation}
% CBB outputs a combination of expected features and mediator features, capturing both generalized and task-specific information.  
% First, the mediator features $\mathcal{M}$ are computed by applying a simple convolutional block to the input features $\mathcal{X}$:  

CBB outputs both expected and mediator features, capturing generalized as well as task-specific information.
The mediator features $\mathcal{M}$ are first obtained by applying a simple convolutional block to the input features $\mathcal{X}_\mathrm{in}$:
\begin{equation}
   \mathcal{M} = \mathrm{Conv}(\mathcal{X}_\mathrm{in}).
\end{equation}

We then estimate the expectations $\hat{\mathbb{E}}[\mathcal{X}]$ (ideally, $\mathbb{E}[\mathcal{X}'] = \mathbb{E}[\mathcal{X}])$ and $\hat{\mathbb{E}}[\mathcal{M}]$ 
using the coefficient estimation method described in Sec.~\ref{sec:coefficients_estimation}. 
The final output is obtained by combining these estimates with the mediator features:

\begin{equation}
   \mathcal{F}_{\text{out}} = \hat{\mathbb{E}}[\mathcal{X}] + \hat{\mathbb{E}}[\mathcal{M}] + \mathcal{M}.
\end{equation}

% The first two terms implement the front-door adjustment, mitigating spurious correlations, while the inclusion of $\mathcal{M}$ preserves task-specific information for downstream processing. Note that the CBB is trained end-to-end using the downstream task loss, with no additional supervision required.

The first two terms implement the front-door adjustment to mitigate spurious correlations, while the inclusion of $\mathcal{M}$ preserves task-specific information for downstream processing. Note that CBB is trained end-to-end with the downstream task loss, without additional supervision.

% What about the loss functions?

%% file: sec/4_experiments.tex
\section{Experiments}
\label{sec:experiments}

\subsection{Datasets}
\noindent\textbf{\textit{Cross-Camera}}: We train on Cityscapes~\cite{cordts2016cityscapes} (2,975 imgs) and evaluate on BDD100K~\cite{yu2020bdd100k}, reporting results on the seven categories shared by both datasets, following~\cite{saito2019strong}.

\noindent\textbf{\textit{Adverse Weather}}: We train on Cityscapes~\cite{cordts2016cityscapes} and evaluate on FoggyCityscapes~\cite{sakaridis2018semantic}, using the most challenging attenuation coefficient split (0.02) to assess robustness.

\noindent\textbf{\textit{Real-to-Artistic}}: We train on VOC~\cite{everingham2010pascal} (16,551 real imgs) and evaluate on three stylized datasets~\cite{inoue2018cross}: Clipart (1K imgs, 20 categories), Comic (2K imgs, 6 categories), and Watercolor (2K imgs, 6 categories), following~\cite{li2022cross}.

\noindent\textbf{\textit{Diverse Weather Datasets}}: We train on Daytime-Sunny (19,395 imgs) and evaluate on Night-Sunny (26,158 imgs), Night-Rainy (2,494 imgs), Dusk-Rainy (3,501 imgs), and Daytime-Foggy (3,775 imgs), following~\cite{wu2022single, lee2024object}.

% \noindent\textbf{\textit{Corruption Cityscapes}}: A test-only benchmark~\cite{michaelis2019benchmarking} evaluating 15 corruption types across five severity levels. We follow the OADG~\cite{lee2024object} settings for comparison.

\noindent\textbf{\textit{Diverse Weather DroneVehicle}}:
Based on the DroneVehicle dataset~\cite{sun2022drone}, we construct one training set and three challenging testing scenarios: \textit{Clear} (8,881 imgs), \textit{Foggy} (1,040 imgs), \textit{Dark} (13,553 imgs), and \textit{Extreme-Dark} (4,965 imgs), corresponding to different levels of visibility and illumination degradation. This extension complements existing UAV detection benchmarks by introducing diverse weather conditions, 
enabling the evaluation of domain generalization under adverse environments.
Models are trained on clear conditions and evaluated on three subsets to 
assess robustness under adverse weather and low-light environments.
More details of this dataset are in the supplementary materials.

\subsection{Implementation Details}

Our experimental setup mostly follows that of Boost~\cite{ he2025boosting} and is implemented based on MMDetection~\cite{chen2019mmdetection}. For all experiments, we use Faster R-CNN~\cite{ren2016faster} as the detection framework. For generative VFMs, the detector is equipped with an FPN~\cite{lin2017feature} and a latent diffusion model~\cite{rombach2022high} as the backbone feature extractor, using frozen weights from Stable Diffusion (SD) v2.1 released by StabilityAI. For discriminative VFMs, we also use Faster R-CNN with backbones initialized from the frozen pretrained weights of DINOv2-L, DINOv3-L, and SAM-Huge. To reduce the computational cost of diffusion-based features, we follow Boost~\cite{he2025boosting} and apply a knowledge distillation strategy~\cite{wang2024crosskd}, where the diffusion-based detector acts as the teacher to guide a smaller student detector. The student model is a Faster R-CNN equipped with a ResNet-101~\cite{he2016deep} backbone. All detectors are trained with SGD using a momentum of 0.9, weight decay of 0.0001, and an initial learning rate of 0.01. Detailed implementation and model structures are provided in the supplementary material.

% BDD100K: ~ref{tab:bdd100k}

% Our experimental setup mostly follows that of GDD~\cite{he2025generalized}. We implement our framework based on MMDetection~\cite{chen2019mmdetection}.
% For generative VFMs, we adopt Faster R-CNN~\cite{girshick2015fast} with an FPN~\cite{lin2017feature} as the detector, and employ a latent diffusion model~\cite{rombach2021highresolution} as the backbone to serve as a feature extractor. The diffusion backbone uses frozen weights from Stable Diffusion (SD) v2.1 released by StabilityAI. Similar to GDD, we apply a knowledge distillation strategy~\cite{wang2024crosskd}, where the diffusion-based detector serves as the teacher to guide a smaller student detector, thereby alleviating the high computational cost of diffusion feature extraction. The student model is a Faster R-CNN equipped with a ResNet-101~\cite{he2016deep} backbone.
% For discriminative VFMs, we use the frozen pretrained weights of the DINOv2-L, DINOv3-L, and SAM-Huge models.
% We train all detectors using the SGD optimizer with a momentum of 0.9, a weight decay of 0.0001, and an initial learning rate of 0.01.
% For a more detailed description of the implementation and model structure, please refer to our supplementary material.

\input{charts/BDD}
\input{charts/foggy}
\input{charts/voc}
\input{charts/drone.tex}
\input{charts/dwd}

% Our experimental setup mostly follows that of GDD~\cite{he2025generalized}. We implement our framework based on the MMDetection~\cite{chen2019mmdetection}. 
% Specifically, we adopt Faster R-CNN~\cite{girshick2015fast} with an FPN~\cite{lin2017feature} as the detector, while employing a latent diffusion model~\cite{rombach2021highresolution} as the backbone to serve as a feature extractor. 
% The diffusion backbone uses frozen weights from Stable Diffusion v2.1 released by StabilityAI.
% Similar to GDD, 
% we apply a knowledge distillation strategy, where a diffusion-based detector serves as the teacher to guide a smaller student detector, thereby alleviating the high computational cost of diffusion feature extraction.
% The student model is a Faster R-CNN equipped with a ResNet-101~\cite{he2016deep} backbone.
% We use the SGD optimizer with a momentum of 0.9 and a weight decay of 0.0001 in the initial learning rate of 0.01.

\subsection{Comparison with State-of-the-Arts}
We present quantitative results on various DG benchmarks in Tables \ref{tab:bdd100k}, \ref{tab:FoggyCityscapes}, \ref{tab:voc}, \ref{tab:drone}, and \ref{tab:dwd}, with AP$_{50}$ as the evaluation metric. 
The following abbreviations are used in the tables: 
\textbf{Com.} (Comic), \textbf{Wat.} (Watercolor), and \textbf{Cli.} (Clipart) for Table~\ref{tab:voc};
\textbf{DF} (Daytime-Foggy), \textbf{DR} (Dusk-Rainy), \textbf{NR} (Night-Rainy), and \textbf{NS} (Night-Sunny) for Table~\ref{tab:dwd};
and \textbf{Ext.~Dark} (Extreme Dark) for Table~\ref{tab:drone}.
To facilitate comparison, \textbf{\textit{Bridge}}’s performance is highlighted in \colorbox{lightgreen}{light green} in the tables.

\noindent\textbf{\textit{Cross-Camera}}:
In Table~\ref{tab:bdd100k}, we evaluate \textbf{\textit{Bridge}} on the Cross-Camera DG benchmark, where the model is trained on Cityscapes and tested on BDD100K.  
\textbf{\textit{Bridge}}, built upon a Stable Diffusion backbone (\textit{Diff.~Detector}), surpasses the previous state-of-the-art method Boost~\cite{he2025boosting} by a margin of $\mathbf{3.8}$ mAP. In addition, following the settings of Boost and GDD, we adopt the CrossKD~\cite{wang2024crosskd} technique to transfer knowledge from Stable Diffusion to a smaller backbone (\textit{Diff.~Guided, R101}), enabling lightweight deployment.  
Compared with Boost’s \textit{Diff.~Guided} detector, our model still achieves a $\mathbf{2.0}$ mAP improvement, further demonstrating that \textbf{\textit{Bridge}} can effectively exploit VFM priors for robust domain generalization, even with a compact backbone.

In terms of discriminative VFMs, \textbf{\textit{Bridge}} achieves consistent mAP improvements of $\mathbf{5.1}$, $\mathbf{1.1}$, and $\mathbf{2.0}$ over the corresponding baselines on DINOv2, DINOv3, and SAM, respectively, further demonstrating its strong generalization capability.  
Notably, we observe substantial improvements in \textit{rider}, \textit{bike}, \textit{person}, and \textit{motor}.  
These categories frequently {co-occur} in real-world driving scenes and exhibit significant {variations} in appearance, pose, and context.  
The consistent gains on these challenging and correlated classes further validate the effectiveness of our proposed CBB, which enables \textbf{\textit{Bridge}} to disentangle causal visual factors and improve generalization under complex domain shifts.

\noindent\textbf{\textit{Adverse Weather}}:
Table~\ref{tab:FoggyCityscapes} presents results on the Adverse Weather DG benchmark, where the model is trained on Cityscapes and evaluated on FoggyCityscapes. Across different backbones, \textbf{\textit{Bridge}} achieves consistent mAP improvements of $\mathbf{5.4}$ for DINOv2, $\mathbf{3.9}$ for DINOv3, and $\mathbf{4.1}$ for SAM, highlighting its backbone-agnostic generalization capability. For diffusion-based detectors, \textbf{\textit{Bridge}} also outperforms Boost and GDD, further demonstrating its effectiveness across diverse detection paradigms.

\noindent\textbf{\textit{Real-to-Artistic}}:
Table~\ref{tab:voc} presents the results on the Real-to-Artistic DG benchmark, where the model is trained on VOC and evaluated on three stylized datasets~\cite{inoue2018cross}: \textit{Clipart}, \textit{Comic}, and \textit{Watercolor}.
Although \textbf{\textit{Bridge}} shows only a slight improvement ($\mathbf{0.4}$ mAP) over \textit{Boost} on the SD backbone, the \textit{Diff.~Guided} detector achieves a larger gain of $\mathbf{1.0}$ mAP. This improvement arises because \textbf{\textit{Bridge}} effectively filters out redundant features, making it more conducive to knowledge distillation.
For discriminative VFMs, \textbf{\textit{Bridge}} yields gains of $\mathbf{4.0}$, $\mathbf{0.6}$, and $\mathbf{0.7}$ mAP over DINOv2, DINOv3, and SAM, respectively. Notably, for DINOv3, despite an already strong baseline of $\mathbf{79.9}$ mAP on the challenging \textit{Clipart} dataset, \textbf{\textit{Bridge}} further boosts performance to $\mathbf{81.0}$ mAP.
% Overall, these results demonstrate the effectiveness of \textbf{\textit{Bridge}} in improving the transferability and generalization capability of both diffusion-based and discriminative VFMs.
Overall, these results demonstrate that \textbf{\textit{Bridge}} improves the transferability and generalization capability of both diffusion-based and discriminative VFMs.

\noindent\textbf{\textit{Diverse Weather Datasets}}: Table~\ref{tab:dwd} reports the results on the Diverse Weather Dataset benchmark, where the model is trained on \textit{Daytime-Sunny} and evaluated on four unseen weather conditions: \textit{Daytime-Foggy}, \textit{Dusk-Rainy}, \textit{Night-Rainy}, and \textit{Night-Sunny}. Across different backbones, \textbf{\textit{Bridge}} consistently improves the average accuracy over strong baselines, with mAP gains of $\mathbf{2.4}$ for the Boost's \textit{Diff.~Detector}, $\mathbf{4.0}$ for the DINOv2 backbone, $\mathbf{2.2}$ for the DINOv3 backbone, and $\mathbf{2.5}$ for the SAM backbone. These results indicate that \textbf{\textit{Bridge}} effectively adapts to challenging conditions such as low-visibility fog, rain-induced occlusions, and low-light nighttime scenes, highlighting its robustness to adverse weather and illumination shifts.
% Table~\ref{tab:dwd} reports results on the Diverse Weather domain generalization benchmark, where the model is trained on Daytime-Sunny and evaluated on four unseen weather conditions: Daytime-Foggy, Daytime-Rainy, Night-Rainy, and Night-Snowy. Across backbone, \textit{Bridge} consistently improves average accuracy over strong baselines, with gains of $\textcolor{red}{2.4}$ for Diffusion Detector with Boost, $\textcolor{red}{4.0}$ for DINOv2 backbone, $\textcolor{red}{2.2}$ for DINOv3 backbone, and $\textcolor{red}{2.5}$ for SAM backbone.

% These results indicate that \textit{Bridge} effectively adapts to challenging conditions, including low-visibility fog, rain-induced occlusions, and low-light nighttime scenes, which underscores its robustness to adverse weather and illumination shifts.

\noindent\textbf{\textit{Diverse Weather DroneVehicle}}:
Table~\ref{tab:drone} presents results on the Diverse Weather DroneVehicle benchmark, where the model is trained on clear conditions and evaluated on Foggy, Dark, and Extreme-Dark subsets. Across all backbones, \textbf{\textit{Bridge}} consistently improves the average performance over strong baselines. Notably, under the Extreme-Dark scenario, the Faster R-CNN baseline performs poorly with only $\mathbf{8.1}$ mAP, whereas VFMs equipped with \textbf{\textit{Bridge}}, such as Diff.~Detector, DINOv2, and DINOv3, achieve significantly higher mAP of $\mathbf{24.2}$, $\mathbf{29.8}$, and $\mathbf{34.0}$, respectively, demonstrating their superior capability in extremely low-light conditions. This improvement is particularly important in extreme-dark scenarios, where images suffer from low signal-to-noise ratios. By leveraging the low-rank basis, \textbf{\textit{Bridge}} can focus on causal representations, preserving essential features even under severely degraded illumination. Gains are also observed in other challenging scenarios, e.g., Foggy and Dark, indicating the robustness of \textbf{\textit{Bridge}} across adverse lighting and weather conditions.

\begin{figure*}[t]
    \centering
    \includegraphics[width=1.02\textwidth]{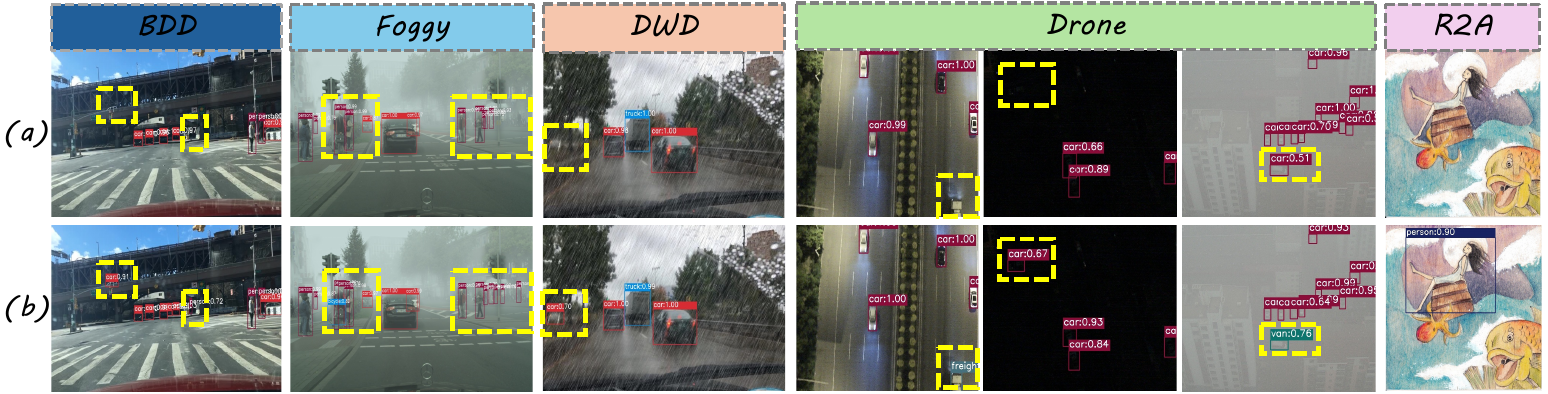}
    \caption{Visualization results on five domain generalization benchmarks: BDD100K (BDD)~\cite{yu2020bdd100k}, FoggyCityscapes (Foggy)~\cite{sakaridis2018semantic}, Diverse Weather Datasets (DWD)~\cite{wu2022single}, Diverse Weather DroneVehicle, and Real-to-Artistic (R2A)~\cite{inoue2018cross}. (a) shows the baseline model with the DINOv3 backbone, while (b) presents the results of our proposed \textbf{\textit{Bridge}} method. Best viewed with zooming in.}
    \label{fig:visualization}
    \vspace{-6pt}
\end{figure*}

\vspace{4pt}
\noindent\textbf{Qualitative results.} 
Figure~\ref{fig:visualization} presents qualitative visualizations using DINOv3 as the backbone detector. \textbf{\textit{Bridge}} produces more robust and precise detections under various challenging scenarios, showing fewer false alarms and missed objects compared to the baseline.

\subsection{Ablation Study}
\noindent{\textbf{Effect of Causal Basis Block}:}
Table~\ref{tab:cbb_component} reports the ablation results of different components within our CBB module. We adopt DINOv3, SAM, and SD as the backbone networks, train them on Cityscapes, and evaluate on Foggy Cityscapes. Compared to the baseline, incorporating Low-Rank Basis  (LRB) leads to mAP improvements of $3.2$, $3.4$, and $1.3$ for DINOv3, SAM, and SD, respectively. Further adding \textit{Sample Queries} for aggregating information from multiple samples provides additional gains of $0.7$, $0.7$, and $0.5$ mAP. These results demonstrate the effectiveness of the low-rank basis and \textit{Sample Queries} components in the proposed CBB module.

We further re-implement the front-door adjustment proposed in GOAT~\cite{wang2024vision} using a cross-attention mechanism~\cite{vaswani2017attention}, which is inserted between multi-scale feature layers (implementation details are provided in the supplementary material). As shown in Table~\ref{tab:facl}, this variant, denoted as FACL, yields only limited improvements across datasets. In contrast, our CBB consistently achieves higher mAP on all benchmarks, highlighting the superior effectiveness of our causal modeling design.

\vspace{4pt}
\noindent{\textbf{Effect of Basis numbers}:}
Table~\ref{tab:basis_ratio} presents the results of varying the number of bases used in our CBB module. 
We adopt DINOv3, SAM, and SD as the backbone networks, train them on Cityscapes, and evaluate on Foggy Cityscapes.
As discussed in Section~\ref{sec:coefficients_estimation}, the basis matrix is defined as $\mathcal{B} = [b_1, \dots, b_K] \in \mathbb{R}^{K \times C}$ with $K < C$, where a smaller $K$ filters out more redundant information and retains more general representations. 
%In this experiment, we investigate how different VFMs respond to varying basis ratios, where the ratio denotes the proportion of $K$ relative to $C$.
In this experiment, we investigate how different VFMs respond to different basis ratios, where the ratio is defined as $K/C$.

We observe that for the strong backbone DINOv3, using only $12.5\%$ of the original feature dimension yields the best performance, suggesting that a compact basis space effectively captures the essential representations. In contrast, SAM and SD achieve optimal results at higher ratios (around $50\%$–$70\%$), indicating that models with weaker representation capacity need a larger basis set to preserve sufficient feature diversity.

\vspace{4pt}

\begin{figure}[t!]
    \centering
    \includegraphics[width=0.46\textwidth]{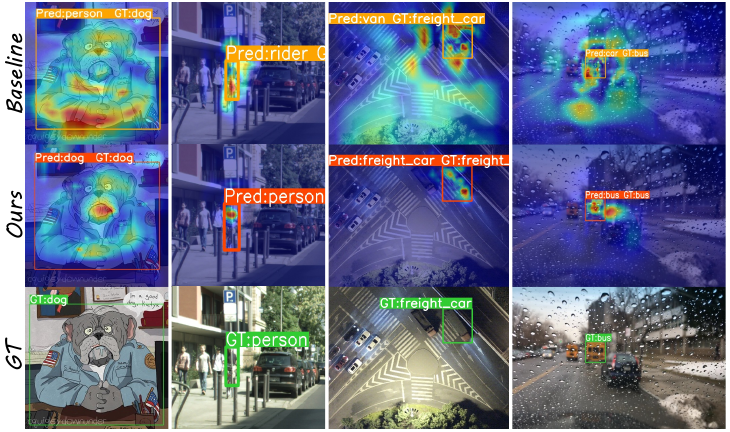}
    \vspace{-3mm}
    \caption{Visualization of the gradient responses of the detector.}
    \label{fig:heatmap}
%\vspace{-mm}
\end{figure}

\noindent\textbf{Empirical Validation of Confounders.}
Figure~\ref{fig:heatmap} visualizes the \textbf{gradient responses} of the detector, revealing the regions used for prediction.
In the \textit{first column} (\textit{real-to-artistic}), the baseline is distracted by the hands and falsely predicts a person, whereas ours focuses on the dog's head.
In the \textit{second column} (\textit{cluttered background}), the baseline is biased by the bike and incorrectly predicts a rider, while ours focuses on the person.
In the \textit{third} and \textit{fourth columns} (\textit{night} and \textit{rainy} scenes), the baseline relies on superficial surrounding cues and produces false predictions, whereas ours attends to critical local regions.

\noindent{\textbf{Generalization across Detectors}:}
Table~\ref{tab:ab_detectors} presents the generalization results of \textbf{\textit{Bridge}} across different object detectors.
In addition to the anchor-based Faster R-CNN, we further evaluate two representative anchor-free detectors, Sparse R-CNN~\cite{sun2021sparse} and TOOD~\cite{feng2021tood}.
Consistent improvements are observed when integrating \textbf{\textit{Bridge}} into both frameworks, highlighting the strong generalization ability of \textbf{\textit{Bridge}} across diverse detection frameworks.

\input{charts/ablation.tex}
\input{charts/detectors}

\vspace{4pt}
\noindent{\textbf{Limitation}:}
In this work, we propose CBB to approximate expectations for formulating the front-door adjustment (Equation~\ref{equ:front_door_adjustment}). However, the complexity of the representation space makes obtaining closed-form solutions intractable~\cite{wang2024vision,yang2021deconfounded}. Consequently, CBB approximates these expectations using the low-rank bases and sample queries, rather than deriving analytical solutions. 

%% file: charts/bdd.tex
% BDD100K
%%%%%%%%%%%%%%%%%%%%%%%%%%%%%%%%%%%%%%%%%%%%%%%%%%%%%%%%%%%%%%%%%%%%%%%%%%%%%%%%%%%%%%%%%%%%%%%%%%%%%%%%%%%%%
\begin{table}[ht]
    \centering
    \caption{DG Results (\%) on BDD100K. Best values are in \textbf{boldface}.}
    \vspace{-4pt}
    \label{tab:bdd100k}
    \setlength{\tabcolsep}{3pt}
    \resizebox{\columnwidth}{!}{%
    \begin{tabular}{l|ccccccc>{\columncolor{gray!20}}c}
    \toprule
    \textbf{Methods} & \textbf{Bike} & \textbf{Bus} & \textbf{Car} & \textbf{Motor} & \textbf{Psn.} & \textbf{Rider} & \textbf{Truck} & \textbf{mAP} \\
    \midrule
    CDSD~\cite{wu2022single} \textit{\scriptsize{(\textcolor{darkblue}{CVPR'22})}} & 22.9 & 20.5 & 33.8 & 14.7 & 18.5 & 23.6 & 18.2 & 21.7 \\
    SHADE~\cite{shade} \textit{\scriptsize{(\textcolor{darkblue}{ECCV'22})}} & 25.1 & 19.0 & 36.8 & 18.4 & 24.1 & 24.9 & 19.8 & 24.0 \\
    MAD~\cite{mad} \textit{\scriptsize{(\textcolor{darkblue}{CVPR'23})}} & - & - & - & - & - & - & - & 28.0 \\
    SRCD~\cite{srcd} \textit{\scriptsize{(\textcolor{darkblue}{TNNLS'24})}} & 24.8 & 21.5 & 38.7 & 19.0 & 25.7 & 28.4 & 23.1 & 25.9 \\
    DDT \footnotesize{\textit{(SD)}}\normalsize~\cite{he2024diffusion}  \textit{\scriptsize{(\textcolor{darkblue}{MM'24})}} & - & - & - & - & - & - & - & 32.7 \\
    GDD \footnotesize{\textit{(Diff. Detector, SD)}}\normalsize~\cite{he2025generalized}  \textit{\scriptsize{(\textcolor{darkblue}{CVPR'25})}} & 38.9 & 31.0 & 71.5 & 37.6 & 61.5 & 47.0 & 38.5 & 46.6 \\

    Boost \footnotesize{\textit{(Diff. Detector,SD)}}\normalsize~\cite{he2025boosting} \textit{\scriptsize{(\textcolor{darkblue}{ICCV'25})}} & 41.2 & 41.7 & 72.7 & 37.2 & 62.8 & 48.7 & 40.5 & 49.3 \\
    \rowcolor{lightgreen} 
    \textit{\textbf{Ours \footnotesize{(Diff. Detector, SD)}}} & \textbf{44.0} & \textbf{43.5} & \textbf{76.2} & \textbf{45.9} & \textbf{66.2} & \textbf{53.9} & \textbf{42.1} & \textbf{53.1} \\
    \midrule
    GDD \footnotesize{\textit{(Diff. Guided, R101)}}\normalsize~\cite{he2025generalized}  \textit{\scriptsize{(\textcolor{darkblue}{CVPR'25})}} & 38.4 & 33.4 & 72.0 & 38.3 & 60.3 & 47.0 & 35.0 & 46.3 \\
    Boost \footnotesize{\textit{(Diff. Guided, R101)}}\normalsize~\cite{he2025boosting} \textit{\scriptsize{(\textcolor{darkblue}{ICCV'25})}} \normalsize & 39.4 & \textbf{34.1} & 72.2 & 37.4 & 61.3 & 46.9 & 35.7 & 46.7 \\
    \rowcolor{lightgreen} 
    \textit{\textbf{Ours \footnotesize{(Diff. Guided, R101)}}} \normalsize & \textbf{40.5} & 33.7 & \textbf{75.3} & \textbf{38.9} & \textbf{62.0} & \textbf{48.8} & \textbf{35.9} & \textbf{47.9} \\

    \midrule
    SAM & 41.1 & 36.7 & 69.0 & 44.1 & 58.5 & 52.0 & 35.5 & 48.1 \\
    \rowcolor{lightgreen} 
    \textit{\textbf{Ours \footnotesize{(SAM backbone)}}} & \textbf{42.6} & \textbf{39.9} & \textbf{70.4} & \textbf{45.5} & \textbf{61.6} & \textbf{52.5} & \textbf{38.2} & \textbf{50.1} \\
     \midrule
    DINOv2& 41.7 & 47.8 & 70.0 & 47.3 & 56.9 & 53.2 & 45.5 &51.8 \\
    \rowcolor{lightgreen} 
    \textit{\textbf{Ours \footnotesize{(DINOv2 backbone)}}} &\textbf{48.2} &\textbf{48.9} &\textbf{74.5}  & \textbf{52.8} & \textbf{65.0} & \textbf{57.1} & \textbf{52.0} & \textbf{56.9}  \\
    \midrule
    DINOv3 & 48.6 & \textbf{56.8} & 73.4 & 52.9 & 64.4 & 57.6 & 51.0 & 57.8 \\
    \rowcolor{lightgreen} 
    \textit{\textbf{Ours \footnotesize{(DINOv3 backbone)}}} & \textbf{49.5} & 54.2 & \textbf{74.3} & \textbf{55.0} & \textbf{66.5} & \textbf{58.2} & \textbf{54.5} & \textbf{58.9} \\

    \bottomrule
    \end{tabular}%
    }
     \vspace{-3mm}
\end{table}

%% file: charts/foggy.tex
\begin{table}[ht]
    \centering
    \caption{DG Results (\%) on FoggyCityscapes. Best values are in \textbf{boldface}.}
    \vspace{-4pt}
    \label{tab:FoggyCityscapes}
    \setlength{\tabcolsep}{2pt}  % 设置统一的列间距
    \resizebox{\columnwidth}{!}{%
    \begin{tabular}{l|cccccccc>{\columncolor{gray!20}}c}
    \toprule
    \textbf{Methods} & \textbf{Bus} & \textbf{Bike} & \textbf{Car} & \textbf{Motor} & \textbf{Psn.} & \textbf{Rider} & \textbf{Train} & \textbf{Truck} & \textbf{mAP} \\
    \midrule
    DIDN~\cite{didn} \textit{\scriptsize{(\textcolor{darkblue}{CVPR'21})}} & 35.7 & 33.1 & 49.3 & 24.8 & 31.8 & 38.4 & 26.5 & 27.7 & 33.4 \\
    FACT~\cite{fact} \textit{\scriptsize{(\textcolor{darkblue}{CVPR'21})}} & 27.7 & 31.3 & 35.9 & 23.3 & 26.2 & 41.2 & 3.0 & 13.6 & 25.3 \\
    FSDR~\cite{fsdr} \textit{\scriptsize{(\textcolor{darkblue}{CVPR'22})}} & 36.6 & 34.1 & 43.3 & 27.1 & 31.2 & 44.4 & 11.9 & 19.3 & 31.0 \\
    MAD~\cite{mad} \textit{\scriptsize{(\textcolor{darkblue}{CVPR'23})}} & 44.0 & 40.1 & 45.0 & 30.3 & 34.2 & 47.4 & 42.4 & 25.6 & 38.6 \\
    DDT \footnotesize{\textit{(Diff. Detector, SD)}}\normalsize~\cite{he2024diffusion} \textit{\scriptsize{(\textcolor{darkblue}{MM'24})}} & - & - & - & - & - & - & - & - & 36.1 \\
    GDD \footnotesize{\textit{(Diff. Detector, SD)}}\normalsize~\cite{he2025generalized}  \textit{\scriptsize{(\textcolor{darkblue}{CVPR'25})}} & 56.2 & 50.4 & 66.7 & 39.9 & 50.2 & 59.5 & 39.9 & 38.0 & 50.1 \\

    Boost\footnotesize{ \textit{(Diff. Detector, SD)}}~\cite{he2025boosting}  \textit{\scriptsize{(\textcolor{darkblue}{ICCV'25})}}& 53.0 & \textbf{55.4} & 68.1 & 42.1 & 51.0 & 59.9 & 39.4 & 36.4 & 50.7 \\
    %\rowcolor{lightgreen}
    \rowcolor{lightgreen}
    \textbf{\textit{Ours\footnotesize{ (Diff. Detector, SD)}}} & \textbf{58.8} & 52.1 & \textbf{72.0} & \textbf{44.2} & \textbf{53.7} & \textbf{61.0} & \textbf{48.1} &\textbf{38.8} & \textbf{53.6} \\
    \midrule
     GDD \footnotesize{\textit{(Diff. Guided, R101)}}\normalsize~\cite{he2025generalized}  \textit{\scriptsize{(\textcolor{darkblue}{CVPR'25})}} & 53.8 & 54.2 & 67.5 & 45.6 & 52.1 & 60.8 & 53.9 & 32.4 & 52.5 \\
    Boost\footnotesize{ \textit{(Diff. Guided, R101)}}~\cite{he2025boosting}  \textit{\scriptsize{(\textcolor{darkblue}{ICCV'25})}}& 55.3 & \textbf{62.7} & 68.2 & 45.5 & 52.9 & \textbf{62.0} & 48.4 & \textbf{37.4} & 54.1 \\
    \rowcolor{lightgreen}
    \textbf{\textit{Ours\footnotesize{ (Diff. Guided, R101)}}} & \textbf{55.7} & 57.1 & \textbf{72.9} & \textbf{46.2} & \textbf{56.5} & 61.8 & \textbf{50.1} & 37.1 & \textbf{54.7} \\
    \midrule
     SAM & 52.0 & 47.7 & 65.9 & 41.8 & 49.2 & 54.6 & 20.8 & \textbf{34.2} & 45.8 \\
    \rowcolor{lightgreen}
    \textit{\textbf{Ours \footnotesize{(SAM backbone)}}} & \textbf{56.4} & \textbf{51.5} & \textbf{68.3} & \textbf{43.1} & \textbf{52.8} & \textbf{58.3} & \textbf{36.5} & 32.3 & \textbf{49.9} \\
        \midrule
    DINOv2 & 62.1 & 45.3 & 65.5 & 44.4 & 47.6 & 55.0 & 56.3 & 46.4 & 52.8 \\
    \rowcolor{lightgreen}
    \textit{\textbf{Ours \footnotesize{(DINOv2 backbone)}}} & \textbf{69.3} & \textbf{55.1} & \textbf{70.1} & \textbf{49.8} & \textbf{54.3} & \textbf{59.4} & \textbf{58.4} & \textbf{49.1} & \textbf{58.2} \\
    \midrule
    DINOv3 & 66.9 & 55.7 & 72.6 & 48.9 & 56.1 & 62.7 & 53.1 & 45.5 & 57.7 \\
    \rowcolor{lightgreen}
    \textit{\textbf{Ours \footnotesize{(DINOv3 backbone)}}} & \textbf{72.1} & \textbf{58.6} & \textbf{73.2} & \textbf{51.7} & \textbf{58.5} & \textbf{64.6} & \textbf{65.4} & \textbf{48.6} & \textbf{61.6} \\

    \bottomrule
    \end{tabular}%
    }
    \vspace{-6pt}
\end{table}

%% file: charts/voc.tex
% VOC
%%%%%%%%%%%%%%%%%%%%%%%%%%%%%%%%%%%%%%%%%%%%%%%%%%%%%%%%%%%%%%%%%%%%%%%%%%%%%%%%%%%%%%%%%%%%%%%%%%%%%%%%%%%%%
\begin{table}[ht]
    \centering
    \caption{DG Results (\%) on Clipart, Comic, and Watercolor. Best values are in \textbf{boldface}.}
    \vspace{-4pt}
    \label{tab:voc}
    \renewcommand{\arraystretch}{1}
    \setlength{\tabcolsep}{10pt}  % 列间距加大一点
    \resizebox{\columnwidth}{!}{%
    \begin{tabular}{l|ccc|>{\columncolor{gray!20}}c}
    \toprule
    \textbf{Methods} & \textbf{Cli.} & \textbf{Com.} & \textbf{Wat.} & \textbf{Average} \\ 
    \midrule
    Div.~\cite{Diversification} \textit{\scriptsize{(\textcolor{darkblue}{CVPR'24})}} & 33.7 & 25.5 & 52.5 & 37.2 \\
    DivAlign~\cite{Diversification} \textit{\scriptsize{(\textcolor{darkblue}{CVPR'24})}} & 38.9 & 33.2 & 57.4 & 43.2 \\
    DDT\footnotesize{ \textit{(Diff. Detector, SD)}}\normalsize~\cite{he2024diffusion} \textit{\scriptsize{(\textcolor{darkblue}{MM'24})}} & 47.4 & 44.4 & 58.7 & 50.2 \\
    GDD\footnotesize{ \textit{(Diff. Detector, SD)}}\normalsize~\cite{he2025generalized} \textit{\scriptsize{(\textcolor{darkblue}{CVPR'25})}} & 58.3 & 51.9 & 68.4 & 59.5 \\
    Boost\footnotesize{ \textit{(Diff. Detector, SD)}}\normalsize~\cite{he2025boosting} \textit{\scriptsize{(\textcolor{darkblue}{ICCV'25})}}   & \textbf{64.1} & \textbf{55.2} & 69.7 & 63.0 \\
    \rowcolor{lightgreen}
    \textit{\textbf{Ours\footnotesize{ (Diff. Detector, SD)}}} & 63.7 & 54.9 & \textbf{71.7} & \textbf{63.4} \\
    \midrule
    GDD\footnotesize{ \textit{(Diff. Guided, R101)}}\normalsize~\cite{he2025generalized} \textit{\scriptsize{(\textcolor{darkblue}{CVPR'25})}} & 40.8 & 29.7 & 54.2 & 41.6 \\
    Boost\footnotesize{ \textit{(Diff. Guided, R101)}}\normalsize~\cite{he2025boosting} \textit{\scriptsize{(\textcolor{darkblue}{ICCV'25})}}  & 40.5 & 30.0 & \textbf{56.6} & 42.4 \\
    \rowcolor{lightgreen}
    \textit{\textbf{Ours\footnotesize{ (Diff. Guided, R101)}}} & \textbf{41.3} & \textbf{32.9} & 56.1 & \textbf{43.4} \\
    \midrule

    SAM & 36.8 & \textbf{35.8} & 48.2 & 40.3 \\
    \rowcolor{lightgreen}
    \textit{\textbf{Ours\footnotesize{ (SAM backbone)}}} & \textbf{38.0} & 34.2 & \textbf{50.9} & \textbf{41.0} \\
    \midrule
    DINOv2 & 71.9 & 54.1 & 70.2 & 65.4 \\
    \rowcolor{lightgreen}
    \textit{\textbf{Ours\footnotesize{ (DINOv2 backbone)}}} & \textbf{76.5} &\textbf{58.5}  & \textbf{73.3} & \textbf{69.4} \\
    \midrule
    DINOv3 & 79.9 & 63.6 & 74.7 & 72.7 \\
    \rowcolor{lightgreen}
    \textit{\textbf{Ours\footnotesize{ (DINOv3 backbone)}}} & \textbf{81.0} & \textbf{64.1} & \textbf{74.9} & \textbf{73.3} \\
    
    \bottomrule
    \end{tabular}%
    }
\end{table}
%%%%%%%%%%%%%%%%%%%%%%%%%%%%%%%%%%%%%%%%%%%%%%%%%%%%%%%%%%%%%%%%%%%%%%%%%%%%%%%%%%%%%%%%%%%%%%%%%%%%%%%%%%%%%

%% file: charts/drone.tex
%%%%%%%%%%%%%%%%%%%%%%%%%%%%%%%%%%%%%%%%%%%%%%%%%%%%%%%%%%%%%%%%%%%%%%%%%%%%%%%%%%%%%%%%%%%%%%%%%%%%%%%%%%%%%
\begin{table}[ht]
    \centering
    \caption{DG Results (\%) on Diverse Weather (Dark, Foggy, and Extreme Dark). Best values are in \textbf{boldface}.}
    \vspace{-4pt}
    \label{tab:drone}
    \renewcommand{\arraystretch}{1}
    \setlength{\tabcolsep}{8pt}
    \resizebox{\columnwidth}{!}{%
    \begin{tabular}{l|ccc|>{\columncolor{gray!20}}c}
    \toprule
    \textbf{Methods} & \textbf{Dark} & \textbf{Foggy} & \textbf{Ext. Dark} & \textbf{Average} \\ 
    \midrule
    Faster R-CNN  & 33.1  & 54.8 &8.1 &  32.0 \\ 
     GDD \footnotesize{ \textit{(Diff. Detector, SD)}}~\cite{he2025generalized} \textit{\scriptsize{(\textcolor{darkblue}{CVPR'25})}} & {40.1} & {54.8} & {20.4} & {38.4} \\
    Boost\footnotesize{ \textit{(Diff. Detector, SD)}}~\cite{he2025boosting} \textit{\scriptsize{(\textcolor{darkblue}{ICCV'25})}} & {42.0} & \textbf{58.3} & {21.3} & {40.5} \\
    \rowcolor{lightgreen}
    \textit{\textbf{Ours\footnotesize{ (Diff. Detector, SD)}}} & \textbf{45.2} & {56.5} &  \textbf{24.2} & \textbf{42.0} \\

     \midrule
     SAM & 40.4 & 52.4  & 19.9 & 37.6 \\
     \rowcolor{lightgreen}
    \textit{\textbf{Ours\footnotesize{ (SAM backbone)}}} & \textbf{41.1} & \textbf{53.2}  & \textbf{20.7}  & \textbf{38.4} \\
    
    \midrule
    DINOv2 & 47.0 & 53.0  & 26.5 & 42.2 \\
    \rowcolor{lightgreen}
    \textit{\textbf{Ours\footnotesize{ (DINOv2 backbone)}}} & \textbf{48.9} & \textbf{58.2}  & \textbf{29.8}  & \textbf{45.6} \\
    \midrule
    DINOv3 & 50.9 &  56.7 & 33.7 & 47.1 \\
    \rowcolor{lightgreen}
    \textit{\textbf{Ours\footnotesize{ (DINOv3 backbone)}}} & \textbf{51.4} & \textbf{59.9} & \textbf{34.0}  & \textbf{48.4} \\
   
     \bottomrule
    \end{tabular}%
    }
    \vspace{-2mm}
\end{table}

%% file: charts/dwd.tex
\begin{table}[ht]
    \centering
    \caption{DG Results (\%) on Diverse Weather Datasets. Best values are in \textbf{boldface}. $*$Results reimplemented by us using 19,395 training images.}
    \vspace{-4pt}
    \label{tab:dwd}
    \setlength{\tabcolsep}{10pt}
    \resizebox{1\columnwidth}{!}{
    \begin{tabular}{l|cccc|>{\columncolor{gray!20}}c}
    \toprule
    \textbf{Methods} & \textbf{DF} & \textbf{DR} & \textbf{NR} & \textbf{NS} & \textbf{Average} \\ 
    \midrule
    CDSD~\cite{wu2022single} \textit{\scriptsize{(\textcolor{darkblue}{CVPR'22})}} & 33.5 & 28.2 & 16.6 & 36.6 & 28.7 \\
    SHADE~\cite{shade} \textit{\scriptsize{(\textcolor{darkblue}{ECCV'22})}} & 33.4 & 29.5 & 16.8 & 33.9 & 28.4 \\
    CLIPGap~\cite{clip_gap} \textit{\scriptsize{(\textcolor{darkblue}{CVPR'23})}} & 32.0 & 26.0 & 12.4 & 34.4 & 26.2 \\
    SRCD~\cite{srcd} \textit{\scriptsize{(\textcolor{darkblue}{TNNLS'24})}} & 35.9 & 28.8 & 17.0 & 36.7 & 29.6 \\
    G-NAS~\cite{gnas} \textit{\scriptsize{(\textcolor{darkblue}{AAAI'24})}} & 36.4 & 35.1 & 17.4 & 45.0 & 33.5 \\
    PhysAug~\cite{physaug} \textit{\scriptsize{(\textcolor{darkblue}{ArXiv'24})}} & 40.8 & 41.2 & 23.1 & 44.9 & 37.5 \\ 
    OA-DG~\cite{lee2024object} \textit{\scriptsize{(\textcolor{darkblue}{AAAI'24})}} & 38.3 & 33.9 & 16.8 & 38.0 & 31.8 \\
    DivAlign~\cite{Diversification} \textit{\scriptsize{(\textcolor{darkblue}{CVPR'24})}} & 37.2 & 38.1 & 24.1 & 42.5 & 35.5 \\
    UFR~\cite{ufr} \textit{\scriptsize{(\textcolor{darkblue}{CVPR'24})}} & 39.6 & 33.2 & 19.2 & 40.8 & 33.2 \\
    Prompt-D~\cite{prompt-d} \textit{\scriptsize{(\textcolor{darkblue}{CVPR'24})}} & 39.1 & 33.7 & 19.2 & 38.5 & 32.6 \\
    DIDM~\cite{didm} \textit{\scriptsize{(\textcolor{darkblue}{ArXiv'25})}} & 39.3 & 35.4 & 19.2 & 42.0 & 34.0 \\
    Boost$^{*}$~\footnotesize{\textit{(Diff. Detector, SD)}}~\cite{he2025boosting} \textit{\scriptsize{(\textcolor{darkblue}{ICCV'25})}} &45.9 &44.3 & 28.2 &48.8 & 41.8 \\
    \rowcolor{lightgreen}
    \textit{\textbf{Ours (Diff. Detector, SD)}} & \textbf{47.5} & \textbf{47.0} & \textbf{30.2} & \textbf{51.9} & \textbf{44.2} \\
    \midrule
    Boost$^{*}$~\footnotesize{\textit{(Diff. Guided, R101)}}~\cite{he2025boosting} \textit{\scriptsize{(\textcolor{darkblue}{ICCV'25})}} &44.4 & 36.1 & 20.9 & 47.7 & 37.3 \\
    \rowcolor{lightgreen}
    \textit{\textbf{Ours (Diff. Guided, R101)}} &\textbf{44.5} & \textbf{38.4} &\textbf{22.4} & \textbf{48.6} & \textbf{38.5} \\
    \midrule
    SAM & 41.2 & 41.3 & 26.1 & 45.0 & 38.4 \\
    \rowcolor{lightgreen}
    \textit{\textbf{Ours (SAM backbone)}} & \textbf{42.7} & \textbf{44.3} & \textbf{28.6} & \textbf{48.0} & \textbf{40.9} \\
    \midrule
    DINOv2 & 40.7 & 43.8 & 33.8 & 44.7 & 40.8 \\
    \rowcolor{lightgreen}
    \textit{\textbf{Ours (DINOv2 backbone)}} & \textbf{44.0} & \textbf{48.7} & \textbf{37.0} & \textbf{49.3} & \textbf{44.8} \\
    \midrule
    DINOv3 & 46.6 & 50.7 & 45.1 & 52.0 & 48.6 \\
    \rowcolor{lightgreen}
    \textit{\textbf{Ours (DINOv3 backbone)}} & \textbf{48.1} & \textbf{53.8} & \textbf{46.7} & \textbf{54.6} & \textbf{50.8} \\
    \bottomrule
    \end{tabular}}

\vspace{-2mm}
\end{table}

%% file: charts/ablation.tex
    % \vspace{-3mm}
    
    % \begin{table}[htbp]
    %     \centering
    %     \caption{Effect of Low-Rank Basis and Sample Queries within CBB.}
    %     \vspace{-5pt}
    %     \label{tab:component}
    %     \footnotesize
    %     \renewcommand{\arraystretch}{1}
    % \setlength{\tabcolsep}{4pt}  % 
    
    % \resizebox{\columnwidth}{!}{%
    %     \begin{tabularx}{\linewidth}{%
    %     >{\centering\arraybackslash}X % Backbone
    %     |>{\centering\arraybackslash}p{2.cm} % LRR
    %     |>{\centering\arraybackslash}p{2.3cm} % Sample Queries 
    %     |>{\centering\arraybackslash}X % mAP
    %     }
    %     \toprule
    %     \textbf{Backbone} & \textbf{LRB} & \textbf{Sample Queries} & \textbf{mAP} \\
    %     \midrule
    %     \multirow{3}{*}{DINOv3}  &           &   &  57.7        \\
    %                              & \ding{51} &  & 60.9          \\
    %                              & \ding{51} & \ding{51} & \textbf{61.6} \\
    %     \midrule
    %     \multirow{3}{*}{SD}      &           &  & 51.8           \\
    %                              & \ding{51} &  & 53.1          \\
    %                              & \ding{51} & \ding{51} & \textbf{53.6} \\
    %     \midrule
    %     \multirow{3}{*}{SAM}    &           &     & 45.8         \\
    %                             & \ding{51} &  & 49.2           \\
    %                              & \ding{51} & \ding{51} & \textbf{49.9} \\
    %     \bottomrule
    %     \end{tabularx}
    %     }
    %     \label{tab:cbb_componet}
    %     \vspace{-6pt}
    %     \end{table}
\begin{table}[t]
\centering
\caption{Ablation of LRB and \textit{Sample Queries} (SQ) in CBB.}
\label{tab:cbb_component}
\footnotesize
\renewcommand{\arraystretch}{0.95}
\setlength{\tabcolsep}{11pt}
\begin{tabular}{ccccc}
\toprule
\textbf{LRB} & \textbf{SQ} & \textbf{DINOv3} & \textbf{SD} & \textbf{SAM} \\
\midrule
 &  & 57.7 & 51.8 & 45.8 \\
\ding{51} &  & 60.9 & 53.1 & 49.2 \\
\ding{51} & \ding{51} & \textbf{61.6} & \textbf{53.6} & \textbf{49.9} \\
\bottomrule
\end{tabular}
\vspace{-4pt}
\end{table}
\input{charts/fd}

\begin{table}[htbp]
    \centering
    \caption{Ablation Study on Basis Ratio}
    \vspace{-3mm}
    \label{tab:basis_ratio}
    \footnotesize
    \renewcommand{\arraystretch}{1}
    \setlength{\tabcolsep}{6pt}  
    \resizebox{\columnwidth}{!}{%
    \begin{tabularx}{\linewidth}{>{\raggedright\arraybackslash}m{0.15\linewidth}|>{\centering\arraybackslash}X>{\centering\arraybackslash}X>{\centering\arraybackslash}X>{\centering\arraybackslash}X>{\centering\arraybackslash}X}
    \toprule
    \multirow{2}{*}{\textbf{Backbone}} & \multicolumn{5}{c}{\textbf{Basis Ratio}} \\
    \cmidrule(lr){2-6}
     & 90\% & 70\% & 50\% & 25\% & 12.5\% \\
    \midrule
    DINOv3 & 60.9 & 61.1 & 61.3 & 61.3 & \textbf{61.6} \\
    SAM    & 48.5 & 48.7 & \textbf{49.9} & 48.5 & 48.4 \\
    SD     & 52.8 & \textbf{53.6} & 52.5 & 52.5 & 51.9 \\
    \bottomrule
    \end{tabularx}
    }
    \vspace{-6pt}
\end{table}

%% file: charts/fd.tex
\begin{table}[htbp]
\centering
\caption{DG Results (\%) on five benchmarks.}
\vspace{-8pt}
\label{tab:map_five_datasets}
\footnotesize
\begin{tabularx}{\linewidth}{
>{\raggedright\arraybackslash}X |
>{\centering\arraybackslash}p{0.8cm} |
>{\centering\arraybackslash}p{0.8cm} |
>{\centering\arraybackslash}p{0.8cm} |
>{\centering\arraybackslash}p{0.8cm} |
>{\centering\arraybackslash}p{0.8cm}
}
\toprule
\textbf{Method} & \textbf{BDD} & \textbf{Foggy} & \textbf{DWD} & \textbf{Drone} & \textbf{R2A} \\
\midrule
Baseline & 57.8 & 57.7 & 48.6 & 47.1 & 72.7 \\
FACL~\cite{wang2024vision}     & 58.5 & 60.5 & 48.2 &  46.8 & 71.6 \\
Ours      & \textbf{58.9} & \textbf{61.6} & \textbf{50.8} & \textbf{48.4} & \textbf{73.3} \\
\bottomrule

\end{tabularx}
\label{tab:facl}
\vspace{-6pt}
\end{table}

%% file: charts/detectors.tex
\begin{table}[htbp]
\centering
\caption{Generalization comparison across different detectors on five datasets.  Best values are in \textbf{boldface}.}
\vspace{-4pt}
\label{tab:generalization_detectors}
\footnotesize
\begin{tabularx}{\linewidth}{
>{\raggedright\arraybackslash}X |
>{\centering\arraybackslash}p{0.6cm} |
>{\centering\arraybackslash}p{0.6cm} |
>{\centering\arraybackslash}p{0.6cm} |
>{\centering\arraybackslash}p{0.6cm} |
>{\centering\arraybackslash}p{0.6cm}
}
\toprule
\textbf{Detector} & \textbf{BDD} & \textbf{Foggy} & \textbf{DWD} & \textbf{Drone} & \textbf{R2A} \\
\midrule
Sparse R-CNN~\cite{sun2021sparse} &36.3  &43.2  &42.3  &28.7  &60.9  \\
 \rowcolor{lightgreen}
\textit{\textbf{Ours}}     & \textbf{37.6} & \textbf{44.6} &\textbf{45.6}  &\textbf{30.3}  &\textbf{62.1}  \\
\midrule
TOOD~\cite{feng2021tood}         & 56.1 & 56.6  & 47.3 & 46.8 &70.3  \\
 \rowcolor{lightgreen}
\textit{\textbf{Ours}}    & \textbf{59.4} & \textbf{60.8} &\textbf{50.1}  & \textbf{50.3} &\textbf{72.8}  \\
\bottomrule
\end{tabularx}
\label{tab:ab_detectors}
\vspace{-6mm}
\end{table}

%% file: sec/5_conclusion.tex
\section{Conclusion}
\label{sec:conclusion}
In this work, we investigate VFMs-based DGOD, revealing that single-source, limited-data training may risk spurious correlations. This motivates us to design \textbf{\textit{Bridge}}, a basis-driven DGOD framework that blocks confounders through causal inference. 
\textbf{\textit{Bridge}} can be seamlessly integrated with both discriminative and generative VFMs in a plug-and-play mechanism with end-to-end training along with the frozen VFM backbones.
Through extensive experiments on five comprehensive benchmarks spanning ground-level scenes to remote sensing imagery, diverse weather conditions, and visual styles, \textbf{\textit{Bridge}} establishes new state-of-the-art performance, demonstrating superior robustness across different visual domains. 
Also, we believe that our extended weather annotations on an existing UAV dataset offer a novel benchmark for domain generalization, contributing to future research in the remote sensing community.